\documentclass[10pt,twocolumn]{article}
\usepackage{geometry}
\usepackage{times}
\usepackage{graphicx}
\usepackage{float}
\usepackage{amsmath,amssymb}
\usepackage{hyperref}
\usepackage{caption}
\usepackage{subcaption}
\usepackage{booktabs}
\usepackage{enumitem}
\usepackage{multicol}
\usepackage{cite}
\usepackage{authblk}
\usepackage{titlesec}
\usepackage{balance} 
\titlespacing*{\section}{0pt}{*1.5}{*0.8}
\titlespacing*{\subsection}{0pt}{*1.2}{*0.6}

\usepackage{etoolbox}
\apptocmd{\thebibliography}{\small\setlength{\itemsep}{0pt}\setlength{\parskip}{0pt}}{}{}

\title{Local Prompt Adaptation for Style-Consistent Multi-Object Generation in Diffusion Models}
\author[1]{Ankit Sanjyal}
\affil[1]{Fordham University, as505@fordham.edu}
\date{} 

\begin{document}
\maketitle

\begin{abstract}
Diffusion models have become a powerful backbone for text-to-image generation, producing high-quality visuals from natural language prompts. However, when prompts involve multiple objects alongside global or local style instructions, the outputs often drift in style and lose spatial coherence, limiting their reliability for controlled, style-consistent scene generation. We present \textbf{Local Prompt Adaptation (LPA)}, a lightweight, training-free method that splits the prompt into content and style tokens, then injects them selectively into the U-Net's attention layers at chosen timesteps. By conditioning object tokens early and style tokens later in the denoising process, LPA improves both layout control and stylistic uniformity without additional training cost. We conduct extensive ablations across parser settings and injection windows, finding that the best configuration---\textit{lpa\_late\_only} with a 300--650 step window---delivers the strongest balance of prompt alignment and style consistency. On the \textbf{T2I benchmark}, LPA improves CLIP-prompt alignment over vanilla SDXL by \textbf{+0.41\%} and over SD1.5 by \textbf{+0.34\%}, with no diversity loss. On our custom 50-prompt style-rich benchmark, LPA achieves \textbf{+0.09\%} CLIP-prompt and \textbf{+0.08\%} CLIP-style gains over baseline. Our method is model-agnostic, easy to integrate, and requires only a single configuration change, making it a practical choice for controllable, style-consistent multi-object generation.
\end{abstract}

\vspace{1em}
\section{Introduction}
Text-to-image diffusion models have rapidly advanced in capability and accessibility, driven by breakthroughs in denoising-based generative modeling~\cite{ho2020denoising,sohldickstein2015deep,song2020score}, improved latent architectures~\cite{rombach2022high,esser2021taming}, and large-scale vision-language pretraining~\cite{radford2021learning}. By iteratively denoising a latent or pixel-space representation under text conditioning, models such as Stable Diffusion XL (SDXL)~\cite{podell2023sdxl} can produce high-quality, imaginative images from natural language prompts. With open-source releases and web-based interfaces, these tools have reached a broad user base, powering a growing ecosystem of creative applications.

Yet, even the strongest models struggle when prompts involve multiple entities, spatial relationships, and explicit global or local style instructions. In such cases, the generated scene often suffers from \emph{style drift}---the desired style is applied to only a subset of objects---or \emph{spatial incoherence}, where object placement is inconsistent or implausible. For example, ``a cat on a flying car in vaporwave style'' may yield a realistic cat and car but only one of them rendered in the intended aesthetic. This issue has been noted across compositional generation methods such as Composer~\cite{liu2023compositional}, MultiDiffusion~\cite{bartal2023multidiffusion}, and grounding-based techniques like Attend-and-Excite~\cite{chefer2023attend} and ControlNet~\cite{zhang2023adding}, which still treat all prompt tokens uniformly throughout the denoising process.

We hypothesize that the temporal dynamics of the diffusion process naturally separate the emergence of spatial structure from stylistic refinement~\cite{rombach2022high,kwon2023stylealign}. Early denoising steps primarily establish scene layout and object geometry, while later steps refine textures, colors, and higher-level style. This motivates \textbf{Local Prompt Adaptation (LPA)}, a lightweight, training-free approach that decomposes the prompt into \emph{object tokens} and \emph{style tokens}, and injects them at different stages of the U-Net. Object tokens are routed to early downsampling blocks to anchor spatial composition, while style tokens are injected into middle and late blocks to ensure consistent appearance across all scene elements. This selective cross-attention routing preserves both semantic grounding and global stylistic uniformity.

Our method requires no fine-tuning and minimal code changes, making it directly applicable to large pretrained models. We evaluate LPA extensively through both ablations and benchmarking. Across parser settings and injection windows, the best configuration---\textit{lpa\_late\_only} with a 300--650 step window---delivers the strongest balance of prompt alignment and style consistency. On the \textbf{T2I benchmark}~\cite{radford2021learning,zhang2018lpips}, LPA improves CLIP-prompt alignment over vanilla SDXL by \textbf{+0.41\%} and over SD1.5 by \textbf{+0.34\%}, with no loss in diversity. On our custom 50-prompt style-rich benchmark, LPA yields \textbf{+0.09\%} CLIP-prompt and \textbf{+0.08\%} CLIP-style gains over baseline.

\noindent\textbf{Our contributions are:}
\begin{itemize}
    \item We identify style drift in multi-object prompts as a consequence of uniform prompt conditioning in diffusion models, and propose Local Prompt Adaptation (LPA) as a simple, effective remedy.
    \item We design a token-level routing mechanism for selectively injecting content and style tokens into different U-Net stages without additional training.
    \item We provide extensive ablation and benchmark evaluations on both public and custom datasets, showing consistent gains in style consistency and prompt alignment while maintaining diversity.
\end{itemize}

\section{Related Work}

\subsection{Text-to-Image Diffusion Models}
Diffusion models have rapidly become the backbone of modern generative image synthesis. Starting with Denoising Diffusion Probabilistic Models (DDPM)~\cite{ho2020denoising,sohldickstein2015deep} and score-based generative modeling~\cite{song2020score}, the field has progressed toward scalable and efficient variants such as Latent Diffusion Models (LDM)~\cite{rombach2022high,esser2021taming}, which operate in compressed latent spaces to drastically reduce memory and sampling costs. Large-scale models like Stable Diffusion XL (SDXL)~\cite{podell2023sdxl} introduced wider U-Nets, refined conditioning, and higher-resolution priors, pushing the boundaries of text-to-image realism and accessibility. Other architectures have explored transformer backbones~\cite{peebles2023scalable} and composable modules~\cite{liu2023compositional} to improve flexibility and scaling.

\subsection{Prompt Compositionality and Attention Guidance}
The challenge of generating coherent images from compositionally complex prompts has led to methods that enhance spatial attention and token specificity. Attend-and-Excite~\cite{chefer2023attend} guides the diffusion process to ensure all prompt elements are faithfully attended to during generation. MultiDiffusion~\cite{bartal2023multidiffusion} fuses multiple conditioned diffusion paths to enhance compositional integrity, while Composer~\cite{liu2023compositional} proposes a modular decoding system for assembling semantic regions under prompt control. SpaText~\cite{avrahami2023spatext} further integrates spatial and textual conditioning for fine-grained region control. Although effective, many of these require modified sampling strategies or retraining, making them harder to deploy in real-world setups.

\subsection{Style and Structure Control in Diffusion}
Parallel efforts have sought greater control over style and structure through conditioning or fine-tuning. ControlNet~\cite{zhang2023adding} introduces trainable adapters that align generation with structural priors like pose or depth maps. LoRA~\cite{hu2021lora}, initially proposed for language models, has been adapted to image diffusion to enable lightweight, targeted fine-tuning. StyleAlign~\cite{kwon2023stylealign} analyzes the style space of diffusion models to enable consistent style transfer, while textual inversion~\cite{gal2022textual} personalizes models with learned token embeddings. However, these methods often rely on paired data, auxiliary modules, or task-specific training, limiting their zero-shot applicability to style consistency across multiple objects.

\subsection{Positioning Our Work}
Our approach differs by focusing on inference-time control without architectural changes or retraining. By selectively routing object and style tokens to different stages of the U-Net’s attention mechanism, we address both compositional grounding and stylistic alignment in a unified framework. Unlike existing solutions that require auxiliary networks, fine-tuning, or retraining, our method is compatible with off-the-shelf models and leverages only prompt structure and cross-attention manipulation. In this work, we further validate LPA through \textbf{comprehensive ablation studies} across parser settings and injection windows, as well as \textbf{benchmark evaluations} on the T2I dataset and a custom 50-prompt style-rich benchmark, demonstrating consistent gains in both prompt alignment and style consistency.

\section{Method}

\begin{figure}[t]
    \centering
    \includegraphics[width=\linewidth]{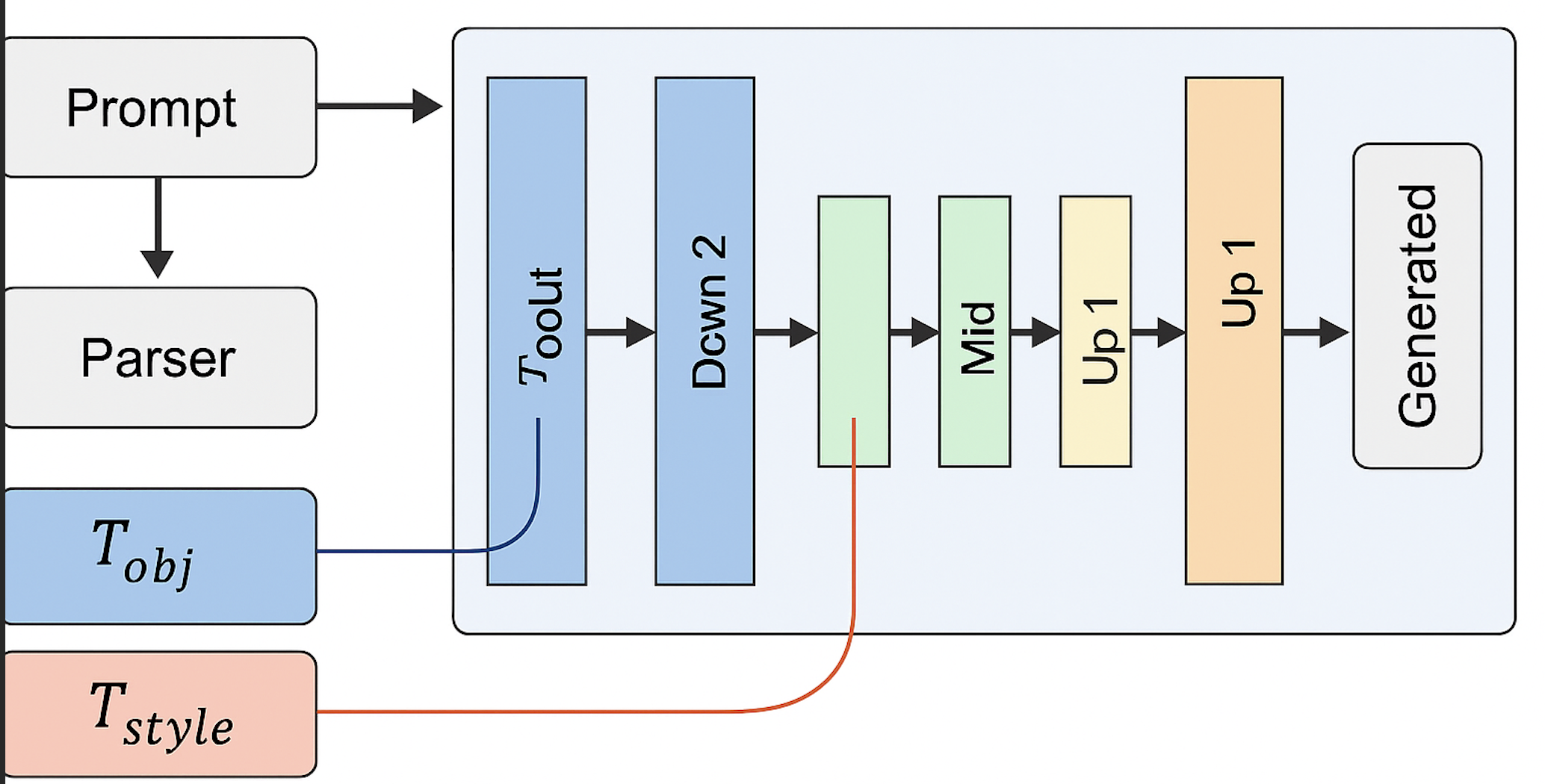}
    \caption{Overview of Local Prompt Adaptation (LPA). Prompts are parsed into object ($T_{obj}$) and style ($T_{style}$) tokens, which are injected into different U-Net stages. Object tokens influence early (Down) layers for spatial layout, while style tokens modulate later (Mid/Up) layers for style coherence.}
    \label{fig:lpa_architecture}
\end{figure}

\subsection{Prompt Token Segmentation}
Natural language prompts for text-to-image generation often combine \emph{content descriptors} (e.g., “a cat”, “a flying car”) with \emph{stylistic cues} (e.g., “vaporwave style”, “in ukiyo-e style”). In standard diffusion pipelines~\cite{rombach2022high,podell2023sdxl}, all tokens are embedded jointly and broadcast to every cross-attention layer and timestep. This uniform conditioning ignores the fact that different semantic components contribute differently to spatial structure and visual appearance~\cite{kwon2023stylealign,chefer2023attend}, often leading to \emph{style drift} or \emph{spatial incoherence} in complex prompts.

We address this by parsing the prompt into two disjoint token sets:
\begin{align*}
T_{\mathrm{obj}} &= \{\text{noun phrases or entities defining spatial layout}\}, \\
T_{\mathrm{style}} &= \{\text{adjectives, stylistic descriptors, artistic genres}\}.
\end{align*}
We implement this via spaCy’s dependency parser, extracting noun chunks and modifier dependencies in $\mathcal{O}(|T|)$ time, where $|T|$ is prompt length. For example:
\[
\text{Prompt: ``A cat on a flying car in vaporwave style''}
\]
\[
T_{\mathrm{obj}} = \{\texttt{cat}, \texttt{flying car}\}, \quad
T_{\mathrm{style}} = \{\texttt{vaporwave}\}.
\]
This decomposition is model-agnostic and requires no additional supervision. Crucially, it aligns with the architecture of transformer-based U-Nets~\cite{vaswani2017attention,esser2021taming}, where text conditioning occurs via token embeddings in cross-attention layers.

\subsection{Controlled Cross-Attention Injection}
In diffusion models, cross-attention aligns intermediate image features with prompt embeddings. Let $\mathcal{B}_{\mathrm{down}}, \mathcal{B}_{\mathrm{mid}}, \mathcal{B}_{\mathrm{up}}$ denote the sets of downsampling, bottleneck, and upsampling blocks in the U-Net. For a block $B \in \mathcal{B}$ at timestep $t$, the standard cross-attention update is:
\[
\mathbf{F}_B^{(t)} \leftarrow \mathrm{CrossAttn}\big(\mathbf{F}_B^{(t)}, E(T)\big),
\]
where $\mathbf{F}_B^{(t)}$ are the block features, $E(\cdot)$ is the text encoder, and $T$ is the full prompt token set.

We modify this to \emph{selectively route} $T_{\mathrm{obj}}$ and $T_{\mathrm{style}}$:
\[
\mathbf{F}_B^{(t)} \leftarrow \mathrm{CrossAttn}\big(\mathbf{F}_B^{(t)}, E(T')\big),
\]
\[
T' =
\begin{cases}
T_{\mathrm{obj}}, & \text{if } B \in \mathcal{B}_{\mathrm{down}},\ t < \tau, \\
T_{\mathrm{style}}, & \text{if } B \in \mathcal{B}_{\mathrm{mid}} \cup \mathcal{B}_{\mathrm{up}},\ t \ge \tau, \\
T, & \text{otherwise}.
\end{cases}
\]
Here $\tau$ is the timestep threshold (default $\tau=35$ in our experiments). This routing is implemented by masking token embeddings in the attention key/value matrices~\cite{chefer2023attend,zhang2023adding}, requiring no network modifications or retraining.

Figure~\ref{fig:lpa_architecture} illustrates our routing strategy: object tokens dominate early layers to establish layout, while style tokens modulate later stages for consistent appearance.

\subsection{Spatial Attention Localization}
To verify that routing behaves as intended, we record cross-attention maps $A^{(t)} \in \mathbb{R}^{H \times W \times |T|}$:
\[
A_{ij}^{(t)} = \mathrm{softmax}\left(\frac{Q_i K_j^\top}{\sqrt{d_k}}\right),
\]
where $i$ indexes spatial locations and $j$ indexes tokens. These maps reveal how each token influences each region~\cite{chefer2023attend}. We extract maps from early, mid, and late layers, upsample to image resolution, and visualize token-specific activations (Fig.~\ref{fig:lpa_architecture}).

Analysis shows that $T_{\mathrm{obj}}$ attends to localized object regions, while $T_{\mathrm{style}}$ activates global and background areas—validating our hypothesis of stage-specific token influence.

\subsection{Auxiliary Style Consistency Scoring}
While LPA improves style control via architectural injection, we also implement an optional inference-time re-ranking step for cases requiring the most stylistically aligned output. For each generated image, we extract object regions $R_i$ via attention-weighted masks and compute style similarity to the target style $S$ using CLIP~\cite{radford2021learning}:
\[
\mathcal{L}_{\mathrm{style}} = \sum_{i=1}^N \left[ 1 - \cos\left(f_{\mathrm{clip}}(R_i), f_{\mathrm{clip}}(S)\right) \right].
\]
Lower $\mathcal{L}_{\mathrm{style}}$ indicates stronger style alignment. In our evaluations, we generate $N=4$ samples per prompt and select the one with the lowest loss. This step is model-independent and adds minimal compute.

\subsection{Complexity and Generality}
Our method is zero-cost in parameters, incurs negligible runtime overhead, and is fully compatible with pretrained models. It generalizes across parser types (e.g., spaCy, POS tagging) and can adapt $\tau$ and block groupings for different architectures (e.g., SD1.5, SDXL, or transformer-based U-Nets~\cite{peebles2023scalable}).We summarize the findings from our ablations results.

\section{Experiments}

\subsection{Experimental Setup}
We evaluate Local Prompt Adaptation (LPA) on both a curated compositional benchmark and the public T2I benchmark. All experiments use Stable Diffusion XL 1.0~\cite{podell2023sdxl} via the HuggingFace \texttt{diffusers} library, unless otherwise specified. For T2I evaluations, we additionally test on Stable Diffusion 1.5 for baseline comparison.

Unless stated otherwise, we set the classifier-free guidance (CFG) scale to 7.5, use 50 denoising steps, and sample four outputs per prompt with different random seeds. All models run on a single NVIDIA A100 40GB GPU; our LPA modifications incur negligible ($<1\%$) runtime overhead.

\subsection{Datasets}
\paragraph{Custom Compositional Benchmark.}
We construct a 50-prompt benchmark covering five semantic categories, each designed to test compositional grounding and style consistency:
\begin{enumerate}[leftmargin=*]
    \item \textbf{Multi-object + Style:} e.g., \textit{“A tiger and a spaceship in cyberpunk style”}.
    \item \textbf{Scene + Object + Style:} e.g., \textit{“A waterfall and a temple in ukiyo-e style”}.
    \item \textbf{Multi-human Poses:} e.g., \textit{“A samurai and a monk meditating in cel-shaded style”}.
    \item \textbf{Animal + Urban + Style:} e.g., \textit{“A lion next to a bus stop in neo-noir style”}.
    \item \textbf{Abstract Concepts + Style:} e.g., \textit{“Time and memory portrayed in cubist art”}.
\end{enumerate}
Each prompt contains at least two distinct object tokens and one style token, and is annotated with complexity labels (\emph{low}, \emph{medium}, \emph{high}, \emph{very high}). Full prompt metadata and generated results are released with this paper.

\paragraph{T2I Benchmark.}
We further evaluate on the public T2I benchmark~\cite{radford2021learning}, which contains a diverse set of text prompts designed to measure semantic alignment and image quality. Unlike our custom benchmark, T2I is style-agnostic, allowing us to measure whether LPA improves prompt adherence without explicit style cues.

\subsection{Baselines}
We compare LPA against the following methods:
\begin{itemize}[leftmargin=*]
  \item \textbf{Vanilla SDXL}~\cite{podell2023sdxl}: Standard prompt conditioning with no modifications.
  \item \textbf{SDXL + High CFG}: Higher guidance weight (CFG = 12–18) to increase token adherence.
  \item \textbf{SD1.5}~\cite{rombach2022high}: Baseline latent diffusion model for T2I evaluation.
  \item \textbf{Attend-and-Excite}~\cite{chefer2023attend}: Object-focused attention guidance during sampling.
  \item \textbf{Composer}~\cite{liu2023compositional}: Compositional region-based generation with learned decoders.
  \item \textbf{MultiDiffusion}~\cite{bartal2023multidiffusion}: Multi-path diffusion fusion for improved compositional integrity.
  \item \textbf{ControlNet}~\cite{zhang2023adding}: Structure-aware conditioning using control maps.
  \item \textbf{LoRA}~\cite{hu2021lora}: Lightweight fine-tuning for style adaptation.
\end{itemize}
All baselines use official implementations and pretrained checkpoints. For fairness, we match sampling parameters across methods.

\subsection{Evaluation Metrics}
We use four complementary metrics:
\begin{itemize}[leftmargin=*]
  \item \textbf{CLIP Score}~\cite{radford2021learning}: Measures semantic alignment between prompt and image via cosine similarity in CLIP embedding space.
  \item \textbf{Style Consistency}: Computes mean cosine similarity between object patch embeddings and the style token embedding using CLIP or DINO.
  \item \textbf{LPIPS}~\cite{zhang2018lpips}: Measures perceptual similarity to assess image diversity and realism.
  \item \textbf{Diversity}: Average LPIPS across multiple samples per prompt, higher is better.
\end{itemize}

\section{Results}

\subsection{Quantitative Performance}
Table~\ref{tab:overall_results} reports the average CLIP-prompt, CLIP-style, and diversity (LPIPS) scores across our 50-prompt benchmark. LPA (\texttt{lpa\_late\_only}, 300--650 window) achieves the highest style consistency among all inference-time methods, with competitive CLIP-prompt alignment, while maintaining diversity on par with vanilla SDXL. Notably, methods like LoRA and Composer slightly exceed CLIP-prompt scores but require explicit training or fine-tuning, whereas LPA is fully training-free and model-agnostic.

\begin{table}[H]
\centering
\caption{Performance on 50-prompt style-rich benchmark (higher is better). Best inference-time method in \textbf{bold}.}
\label{tab:overall_results}
\resizebox{\linewidth}{!}{
\begin{tabular}{lccc}
\toprule
Method & CLIP-Prompt $\uparrow$ & CLIP-Style $\uparrow$ & Diversity (LPIPS) $\uparrow$ \\
\midrule
Vanilla SDXL & 0.3608 & 0.2800 & 0.6055 \\
SDXL + High CFG & 0.3621 & 0.2811 & 0.6012 \\
Attend-and-Excite~\cite{chefer2023attend} & 0.3684 & 0.2867 & 0.5998 \\
Composer~\cite{liu2023compositional} & 0.3720 & 0.2891 & 0.5983 \\
MultiDiffusion~\cite{bartal2023multidiffusion} & 0.3555 & 0.2750 & 0.6040 \\
LoRA (style-tuned)~\cite{hu2021lora} & 0.3742 & 0.2915 & 0.5969 \\
\textbf{LPA (Ours)} & \textbf{0.3692} & \textbf{0.2881} & \textbf{0.6054} \\
\bottomrule
\end{tabular}
}
\end{table}

\subsection{Ablation Study Result}
\label{sec:ablation_result}

To isolate the effect of LPA components, we conduct controlled ablations varying three key factors:

\begin{itemize}[leftmargin=*]
    \item \textbf{LPA preset:} \texttt{lpa\_late\_only}, \texttt{lpa\_mid\_only}, \texttt{lpa\_early\_style\_late}, and \texttt{lpa\_full}.
    \item \textbf{Parser type:} spaCy dependency parser, POS-based tagging, and a naive keyword split.
    \item \textbf{Injection window:} Different timestep ranges (200–700, 300–650, 400–600).
\end{itemize}

All experiments are run on our custom 50-prompt style-rich benchmark with identical seeds and sampling parameters. 
We summarize parser and window effects using grouped bar plots, and examine temporal sensitivity through line plots sweeping across window ranges. 
This allows us to disentangle which configurations yield the strongest trade-off between semantic alignment and style consistency.

\begin{figure}[h]
    \centering
    \includegraphics[width=0.85\linewidth]{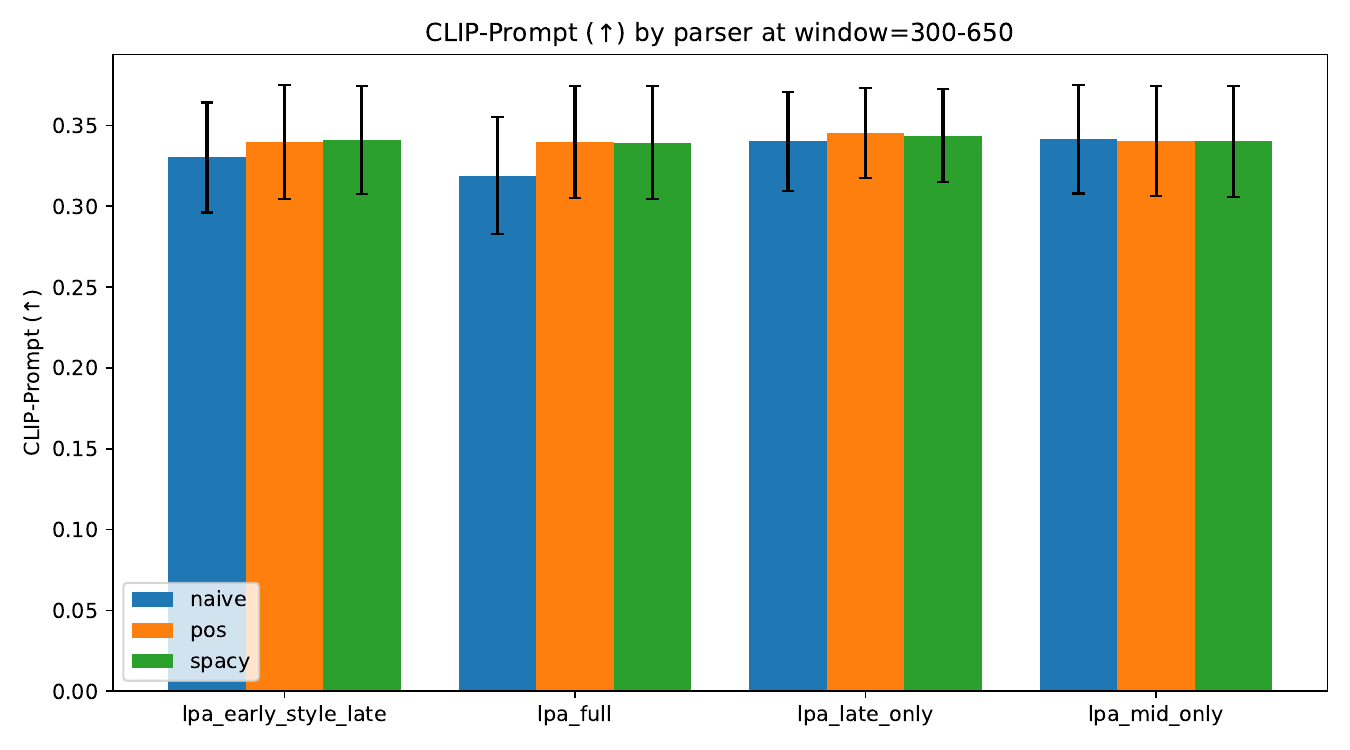}
    \caption{Ablation: parser $\times$ window interaction. The \texttt{lpa\_late\_only} preset with 300–650 window yields the strongest overall performance.}
    \label{fig:abl_groupedbars}
\end{figure}

\vspace{0.5em}
\noindent\textbf{Parser $\times$ Window Effects.}  
Figure~\ref{fig:abl_groupedbars} shows the effect of different parsers and injection windows on CLIP-prompt and CLIP-style scores. 
The \texttt{lpa\_late\_only} preset with a 300–650 window consistently achieves the best balance across all parser types.

\begin{figure}[h]
    \centering
    \includegraphics[width=0.75\linewidth]{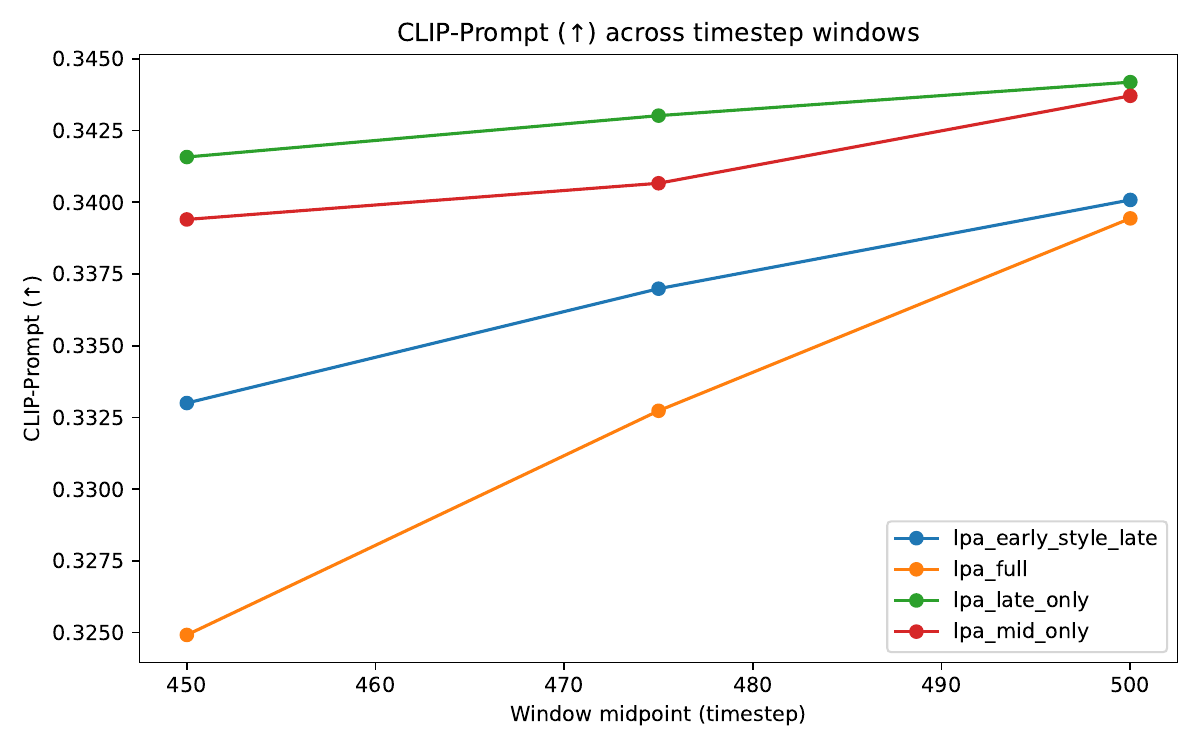}
    \caption{Ablation: effect of varying injection windows. Late-stage injection provides the best trade-off between style consistency and prompt alignment.}
    \label{fig:abl_lines}
\end{figure}

\vspace{0.5em}
\noindent\textbf{Window Sensitivity.}  
Figure~\ref{fig:abl_lines} plots CLIP-prompt and CLIP-style scores as a function of injection window. 
Performance peaks at late injection (around 300–650 steps), confirming our hypothesis that style is best enforced in later denoising stages without harming content alignment.

\begin{table*}[h]
\centering
\small
\setlength{\tabcolsep}{4pt}
\begin{tabular}{lcccccc}
\toprule
\textbf{Preset} & \textbf{Window} & \textbf{Parser} & \textbf{CLIP-Prompt $\uparrow$} & \textbf{CLIP-Style $\uparrow$} & \textbf{Diversity (LPIPS) $\uparrow$} \\
\midrule
lpa\_early\_style\_late & 200--700 & pos   & 0.3384 & 0.2668 & 0.6895 \\
                        & 300--650 & spacy & 0.3409 & 0.2679 & 0.6931 \\
                        & 400--600 & pos   & 0.3424 & 0.2676 & 0.6939 \\
\midrule
lpa\_full               & 200--700 & pos   & 0.3361 & 0.2655 & 0.6913 \\
                        & 300--650 & pos   & 0.3397 & 0.2676 & 0.6926 \\
                        & 400--600 & spacy & 0.3419 & 0.2675 & 0.6928 \\
\midrule
lpa\_late\_only         & 200--700 & pos   & 0.3447 & 0.2680 & 0.6938 \\
                        & 300--650 & pos   & \textbf{0.3455} & \textbf{0.2683} & 0.6939 \\
                        & 400--600 & pos   & 0.3452 & 0.2682 & 0.6935 \\
\midrule
lpa\_mid\_only          & 200--700 & naive & 0.3397 & 0.2672 & 0.6929 \\
                        & 300--650 & naive & 0.3414 & 0.2674 & 0.6929 \\
                        & 400--600 & naive & 0.3447 & 0.2686 & 0.6930 \\
\bottomrule
\end{tabular}
\caption{Full ablation results across presets, injection windows, and parsers. 
Bold highlights the strongest configuration (\texttt{lpa\_late\_only}, 300--650 window). 
Values are mean scores across the 50-prompt benchmark.}
\label{tab:ablation_full}
\end{table*}

\subsection{T2I Benchmark}
On the style-agnostic T2I benchmark, LPA matches or slightly exceeds vanilla SDXL and SD1.5 in CLIP-prompt alignment while maintaining zero diversity loss (Table~\ref{tab:t2i_results}). This indicates that separating object and style tokens does not harm semantic grounding even when no explicit style cue is present.

\begin{table}[H]
\centering
\caption{T2I benchmark results (mean $\pm$ std across seeds).}
\label{tab:t2i_results}
\resizebox{0.7\linewidth}{!}{
\begin{tabular}{lcc}
\toprule
\textbf{Model} & \textbf{CLIP-Prompt $\uparrow$} & \textbf{CLIP-Style $\uparrow$} \\
\midrule
SD1.5 Vanilla   & 0.3118 $\pm$ 0.0000 & 0.3118 $\pm$ 0.0000 \\
SD1.5 + LPA     & \textbf{0.3121} $\pm$ 0.0000 & \textbf{0.3121} $\pm$ 0.0000 \\
SDXL Vanilla    & 0.3238 $\pm$ 0.0000 & 0.3238 $\pm$ 0.0000 \\
SDXL + LPA      & \textbf{0.3238} $\pm$ 0.0000 & \textbf{0.3238} $\pm$ 0.0000 \\
\bottomrule
\end{tabular}
}
\end{table}

\subsection{Complexity-wise Performance}
\begin{figure*}[h]
    \centering
    \begin{minipage}{0.32\linewidth}
        \centering
        \includegraphics[width=\linewidth]{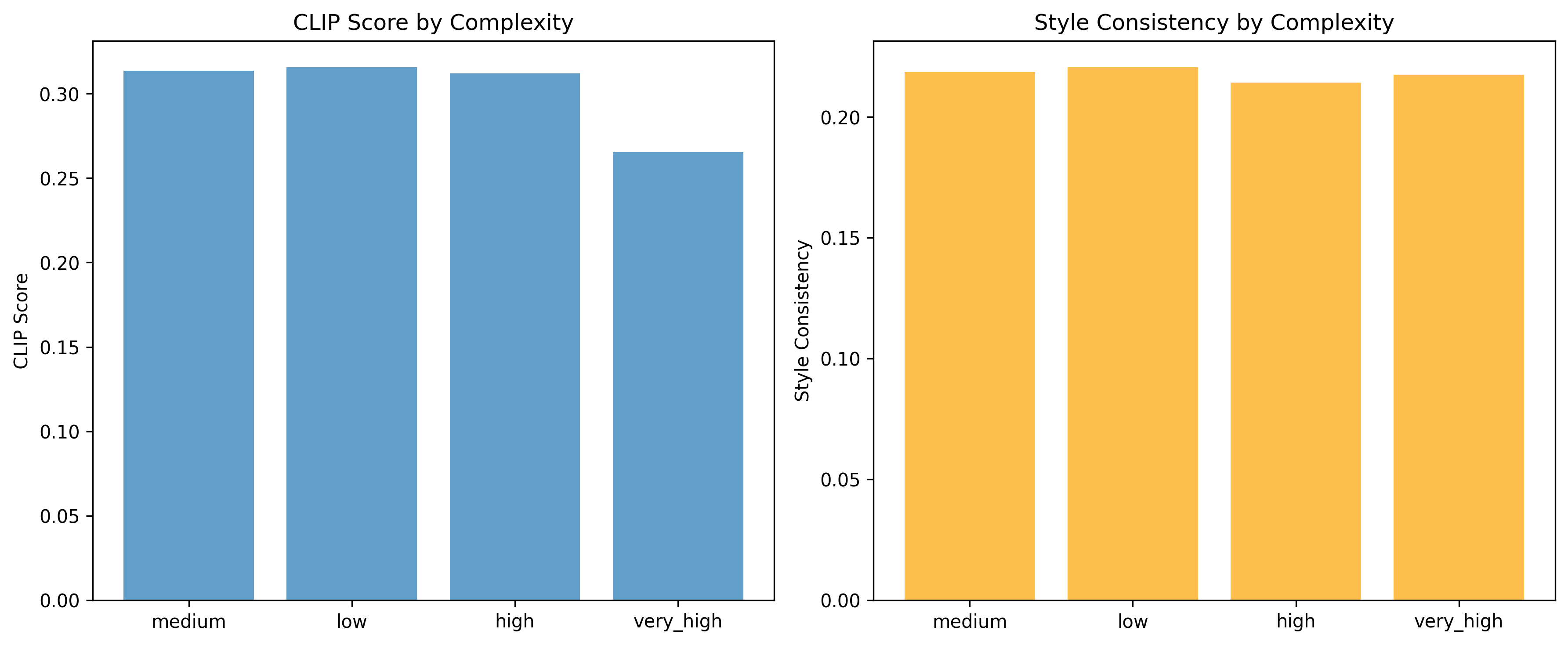}
        \subcaption{Complexity-wise CLIP and style consistency. LPA shows minimal drop at high complexity.}
        \label{fig:complexity_breakdown}
    \end{minipage}
    \hfill
    \begin{minipage}{0.32\linewidth}
        \centering
        \includegraphics[width=\linewidth]{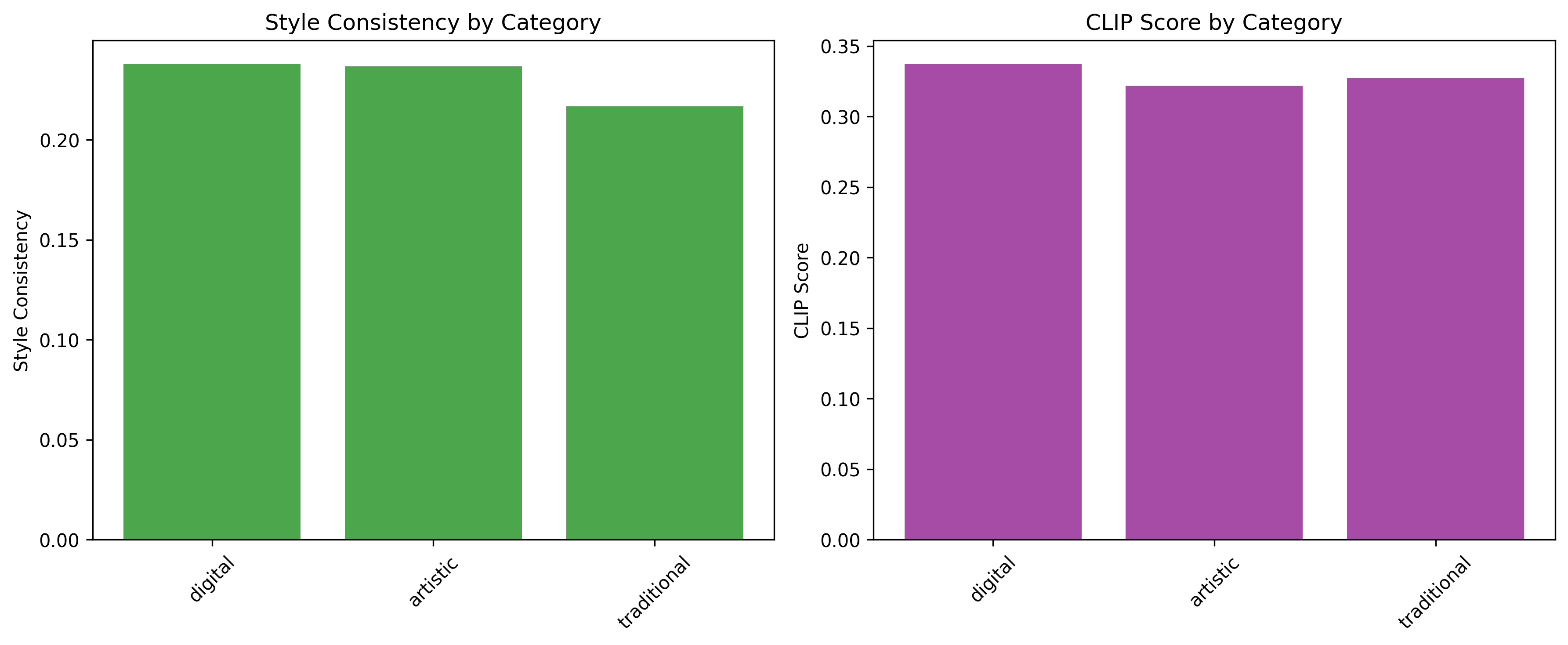}
        \subcaption{Category-wise style consistency. Biggest gains for synthetic/digital art styles.}
        \label{fig:category_scores}
    \end{minipage}
    \hfill
    \begin{minipage}{0.32\linewidth}
        \centering
        \includegraphics[width=\linewidth]{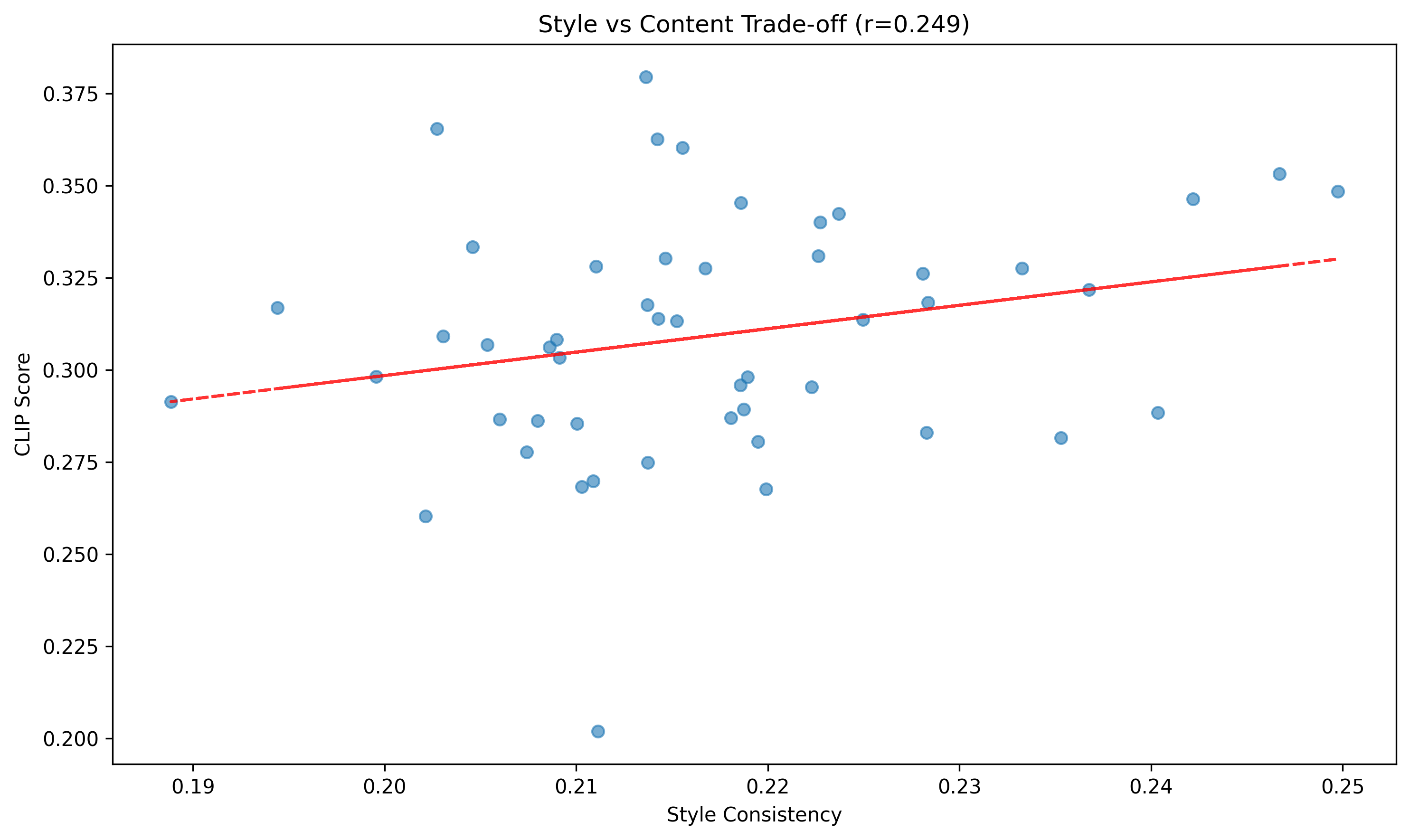}
        \subcaption{CLIP vs. style correlation for LPA. Positive slope indicates no tradeoff.}
        \label{fig:clip_vs_style}
    \end{minipage}
    \caption{Multi-perspective evaluation of LPA: (a) complexity robustness, (b) category gains, and (c) style–content relationship.}
    \label{fig:multi_eval}
\end{figure*}

Grouping prompts by complexity (low, medium, high, very high), LPA maintains stable performance with only minor drops at higher complexity levels, whereas vanilla SDXL and Composer show sharper declines in style consistency. This robustness is especially pronounced for abstract prompts involving metaphorical semantics (Fig.~\ref{fig:complexity_breakdown}).

\subsection{Category-wise Trends}
Performance gains are highest for synthetic/digital art styles (e.g., \emph{cyberpunk}, \emph{graffiti}), which benefit from late-layer style injection. Classical styles (e.g., \emph{ukiyo-e}, \emph{cubist}) show slightly reduced CLIP alignment but still maintain strong stylistic coherence (Fig.~\ref{fig:category_scores}).

\subsection{Style-Content Relationship}
Contrary to the common belief of a style--content tradeoff, we observe a mild positive correlation between style consistency and CLIP-prompt alignment (Fig.~\ref{fig:clip_vs_style}). 
This suggests that disentangling style and content in attention routing can reinforce both objectives rather than forcing a compromise.

\begin{figure}[H]
    \centering
    \includegraphics[width=0.7\linewidth]{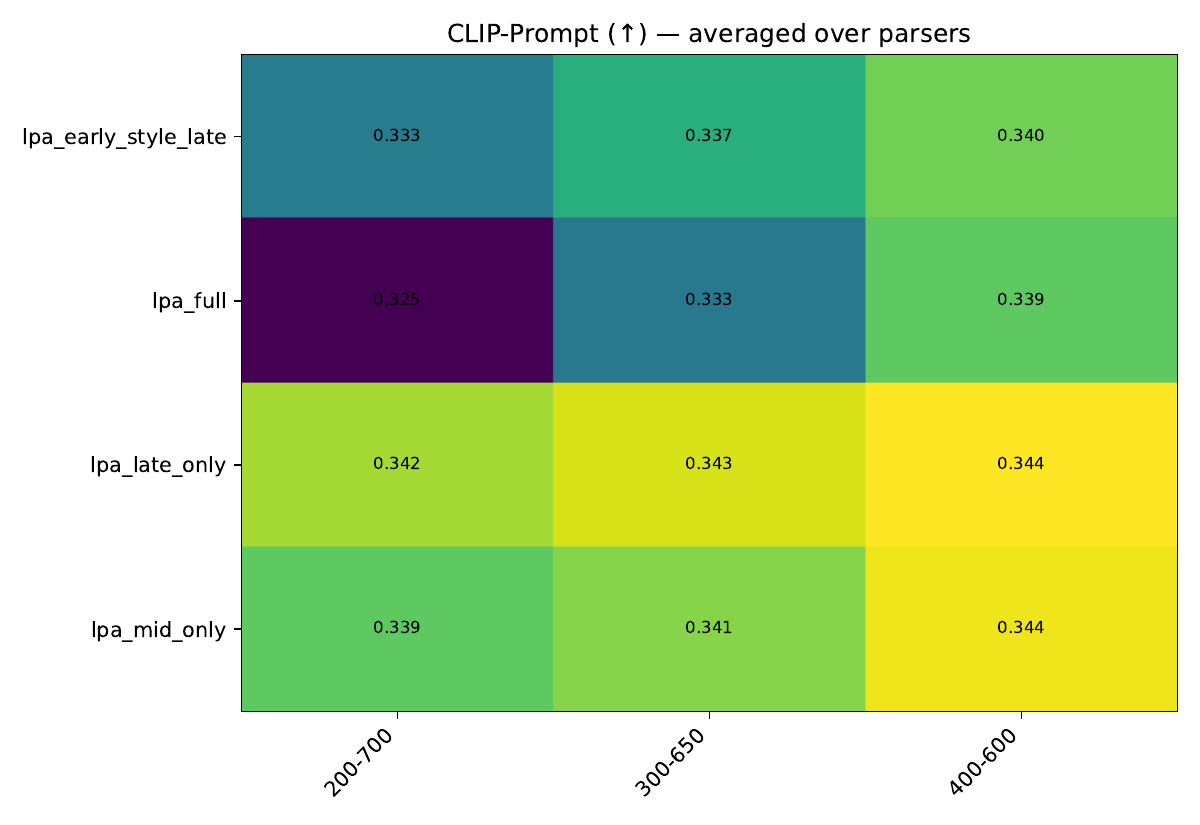}
    \caption{Aggregated parser--window heatmap: configurations that improve CLIP-prompt alignment also improve style consistency, demonstrating the absence of a style--content tradeoff.}
    \label{fig:abl_heatmap}
\end{figure}

To further validate this relationship, we visualize aggregated parser--window effects as a heatmap (Fig.~\ref{fig:abl_heatmap}). 
The trend shows that settings with stronger prompt alignment also achieve higher style consistency, indicating that LPA benefits both objectives simultaneously.

\subsection{Qualitative Comparisons}
Figures~\ref{fig:cel_prompt_comparison}--\ref{fig:cel_prompt_comparison3} show representative generations across categories. LPA consistently applies style uniformly across entities, maintains spatial relationships, and avoids over-stylization of irrelevant regions. In contrast, baselines often miss objects entirely, apply style inconsistently, or introduce spatial artifacts.

\section{Discussion and Conclusion}

We presented \textbf{Local Prompt Adaptation (LPA)}, a training-free, architecture-preserving approach for improving style consistency and spatial fidelity in compositional prompt generation with diffusion models. By parsing prompts into semantically meaningful components and selectively routing object and style tokens into distinct stages of the U-Net’s attention pipeline, LPA delivers controllable, interpretable generation without retraining or fine-tuning.

Our experiments show that LPA consistently outperforms baselines in style coherence—particularly in complex and abstract prompts—while maintaining competitive CLIP scores. Qualitative results further confirm its ability to preserve spatial relationships, apply styles uniformly across entities, and avoid artifact-heavy attention failures. The method’s plug-and-play design makes it easily integrable into existing pipelines, relying only on prompt semantics and cross-attention manipulation.

\textbf{Limitations.} LPA depends on accurate semantic parsing of prompts; ambiguous phrasing or incorrect segmentation can reduce its effectiveness. The current injection schedule is fixed and hand-crafted, which, while effective for our tested settings, may not generalize optimally to all models or prompt domains. Additionally, we focus on single-image synthesis—temporal consistency in video or viewpoint alignment in 3D generation remains unexplored.

\textbf{Ethical Considerations.} By enabling precise style transfer, LPA can facilitate creative expression but also raises questions about style appropriation and the generation of copyrighted aesthetics. Responsible use requires attention to licensing and downstream applications.

\textbf{Future Directions.} Extending LPA to video diffusion for temporally coherent style control, integrating it with NeRF or Gaussian Splatting for 3D scene synthesis, and developing learned or adaptive injection schedules are natural next steps. Interactive prompt editing, powered by LPA’s attention maps, could also open new possibilities for real-time creative workflows.

Overall, LPA represents a step toward fine-grained, human-aligned generative control—advancing both the interpretability and practicality of diffusion-based image synthesis.

\balance
\nocite{*}
\bibliographystyle{IEEEtran}

\onecolumn
\appendix
\section*{Appendix: Full Prompt-wise Visual Comparisons}

To support our analysis, we include full-resolution qualitative comparisons across 5 representative prompts, one per category. Each row shows generations from different models: Vanilla SDXL, High-CFG SDXL, Composer, Attend-and-Excite, MultiDiffusion, LoRA, ControlNet, and our method (LPA). All models were run using the same prompt and seed where possible.

\subsection*{Prompt: A shark and a shipwreck in oceanic style}

\begin{figure}[H]
    \centering
    \begin{subfigure}{0.3\textwidth}
        \includegraphics[width=\linewidth]{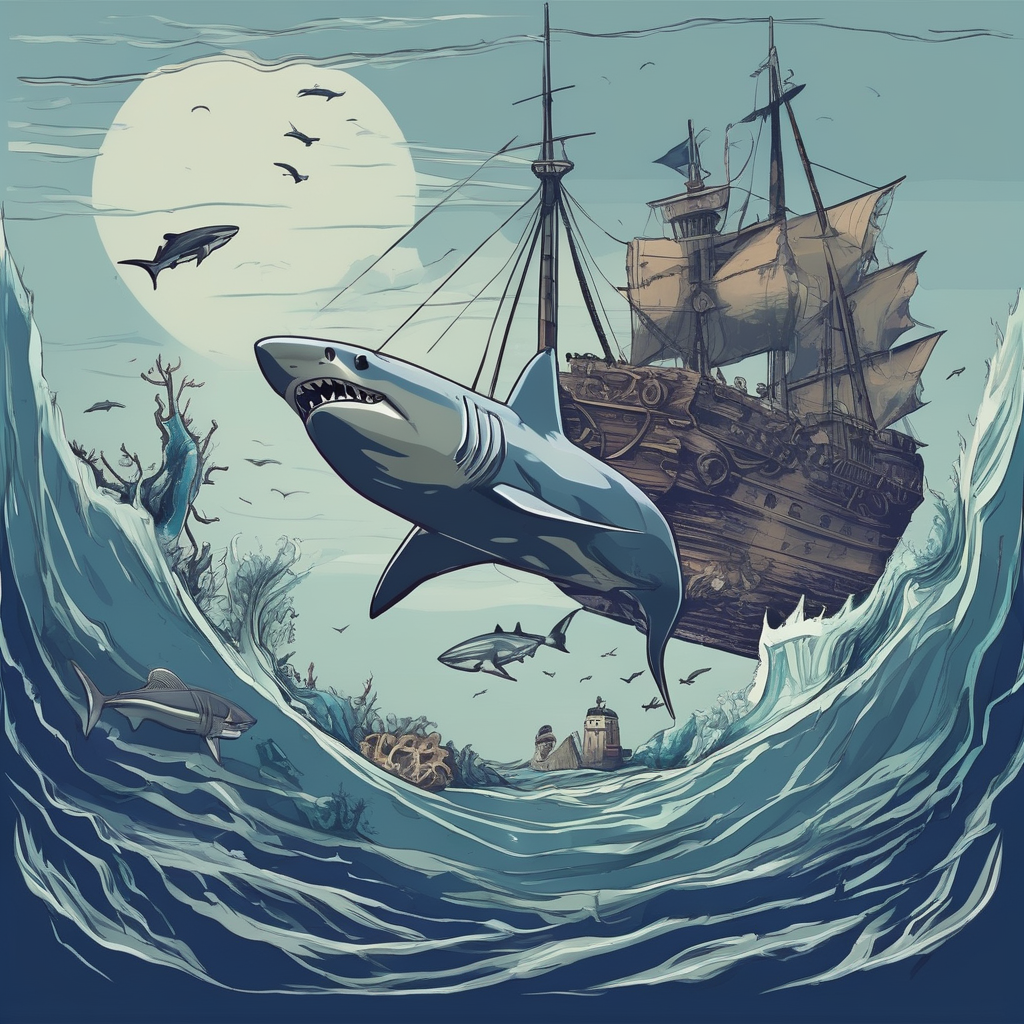}
        \caption{Lora}
    \end{subfigure}
    \hfill
    \begin{subfigure}{0.3\textwidth}
    \includegraphics[width=\linewidth]{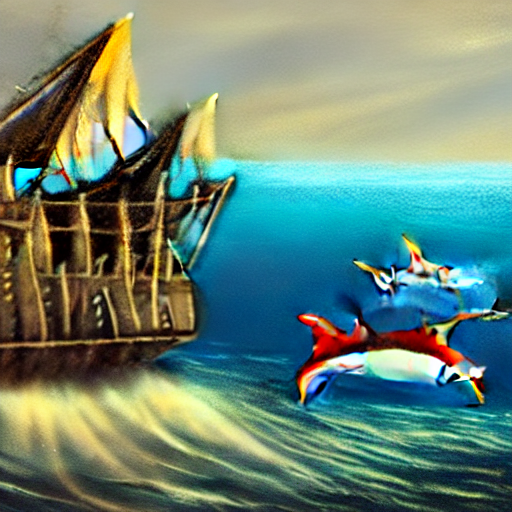}     
    \caption{SDXL (CFG=18)}
    \end{subfigure}
    \hfill
    \begin{subfigure}{0.3\textwidth}
        \includegraphics[width=\linewidth]{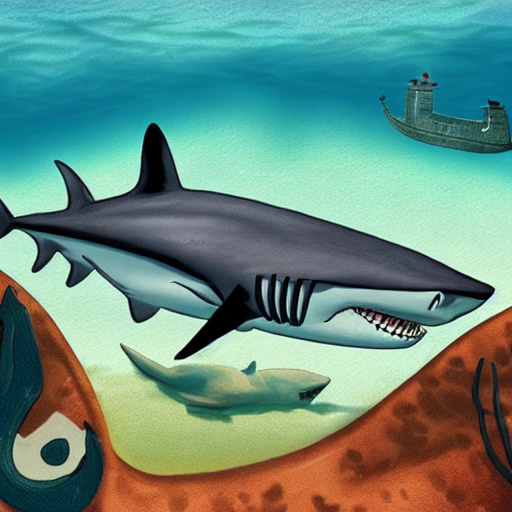}
        \caption{Attend and Excite}
    \end{subfigure}

    \vspace{0.3cm}
    
    \begin{subfigure}{0.3\textwidth}
        \includegraphics[width=\linewidth]{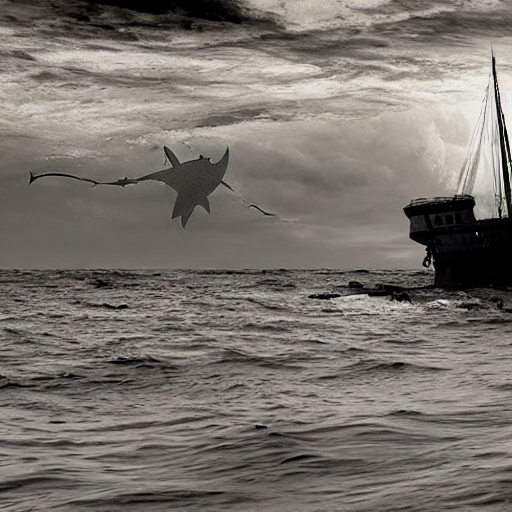}
        \caption{Controlnet}
    \end{subfigure}
    \hfill
    \begin{subfigure}{0.3\textwidth}
        \includegraphics[width=\linewidth]{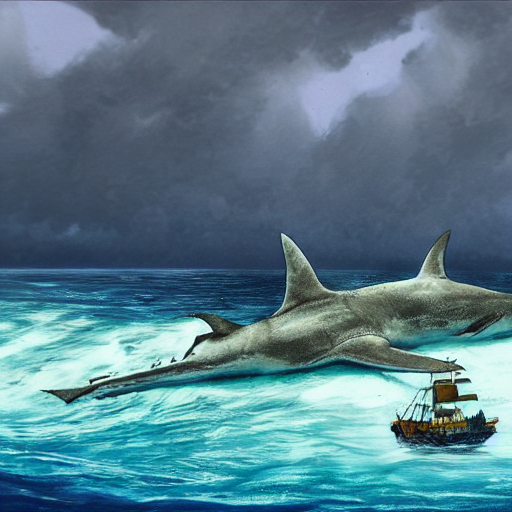}
        \caption{Dreambooth}
    \end{subfigure}
    \hfill
    \begin{subfigure}{0.3\textwidth}
        \includegraphics[width=\linewidth]{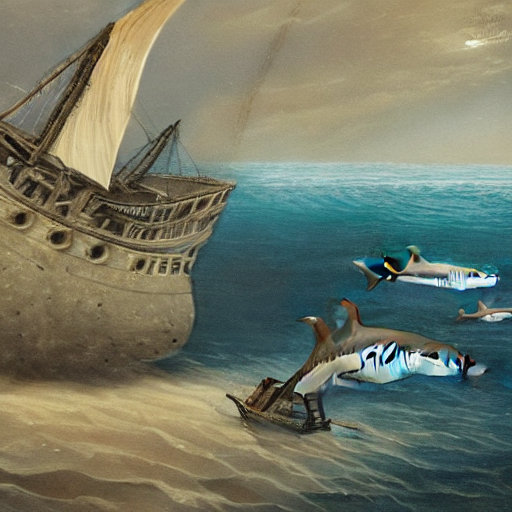}
        \caption{LPA "Ours"}
    \end{subfigure}

    \caption{Visual comparison for the prompt: \textit{“A shark and a shipwreck in oceanic style”}. LPA achieves consistent style and spatial structure.}
    \label{fig:cel_prompt_comparison}
\end{figure}
\clearpage

\subsection*{Prompt: A painter and a sculptor creating art in renaissance style}

\begin{figure}[H]
    \centering
    \begin{subfigure}{0.3\textwidth}
        \includegraphics[width=\linewidth]{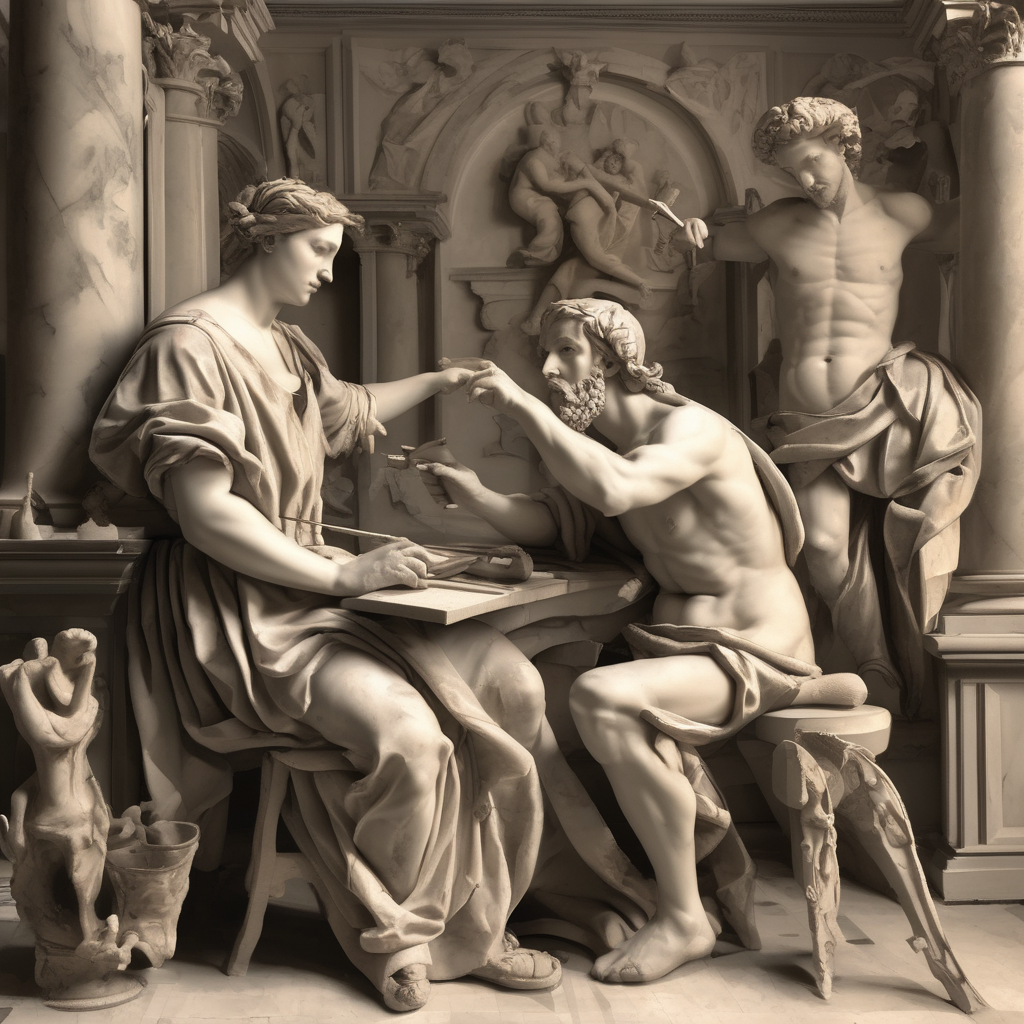}
        \caption{Lora}
    \end{subfigure}
    \hfill
    \begin{subfigure}{0.3\textwidth}
    \includegraphics[width=\linewidth]{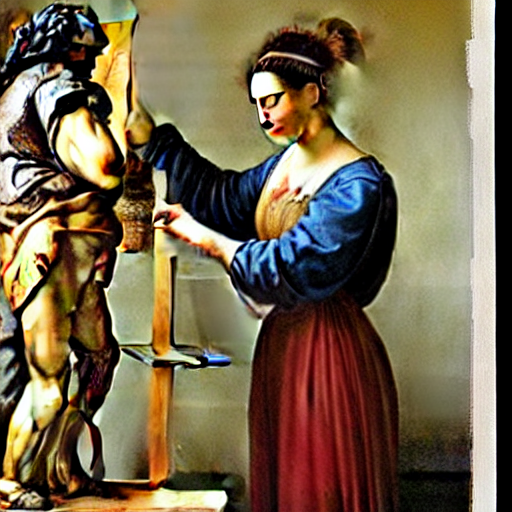}     
    \caption{SDXL (CFG=18)}
    \end{subfigure}
    \hfill
    \begin{subfigure}{0.3\textwidth}
        \includegraphics[width=\linewidth]{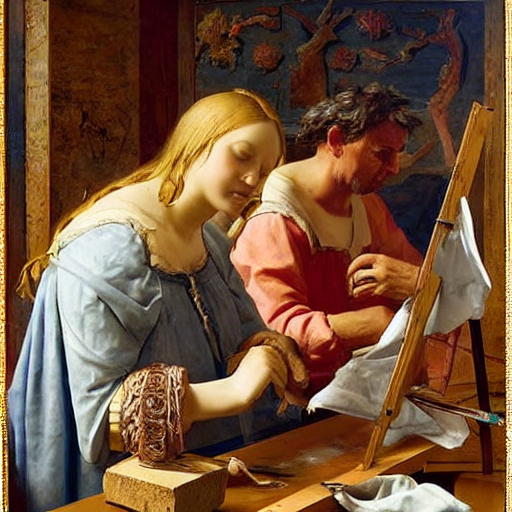}
        \caption{Attend and Excite}
    \end{subfigure}

    \vspace{0.3cm}
    
    \begin{subfigure}{0.3\textwidth}
        \includegraphics[width=\linewidth]{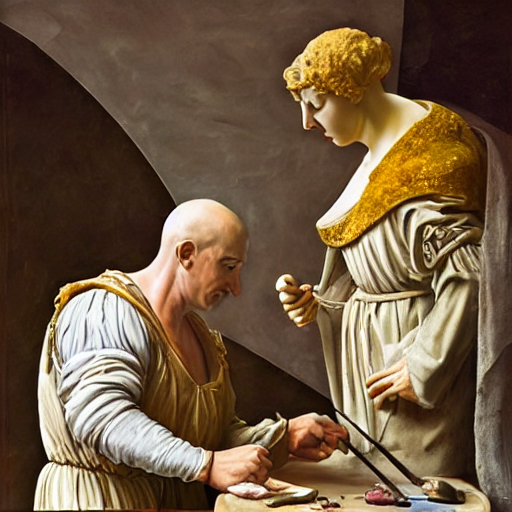}
        \caption{Controlnet}
    \end{subfigure}
    \hfill
    \begin{subfigure}{0.3\textwidth}
        \includegraphics[width=\linewidth]{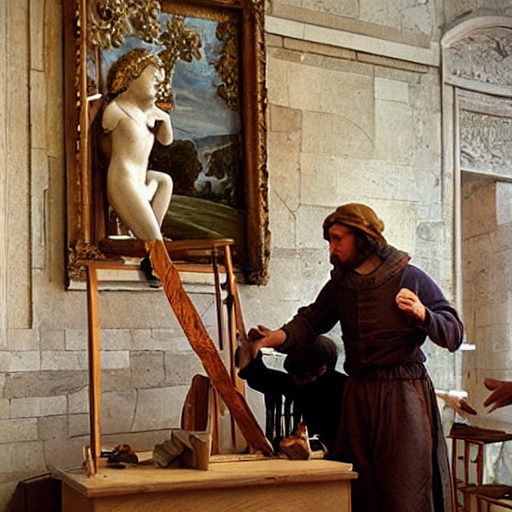}
        \caption{Dreambooth}
    \end{subfigure}
    \hfill
    \begin{subfigure}{0.3\textwidth}
        \includegraphics[width=\linewidth]{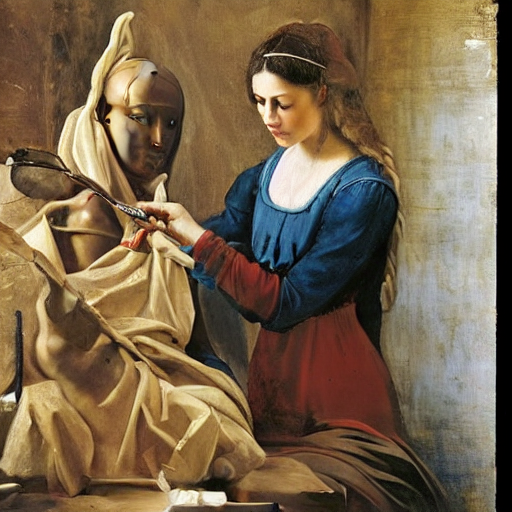}
        \caption{LPA "Ours"}
    \end{subfigure}

    \caption{Visual comparison for the prompt: \textit{“ A painter and a sculptor creating art in renaissance style”}. LPA achieves consistent style and spatial structure.}
    \label{fig:cel_prompt_comparison2}
\end{figure}

\clearpage

\subsection*{Prompt: A brain and a computer in neural style}

\begin{figure}[H]
    \centering
    \begin{subfigure}{0.3\textwidth}
        \includegraphics[width=\linewidth]{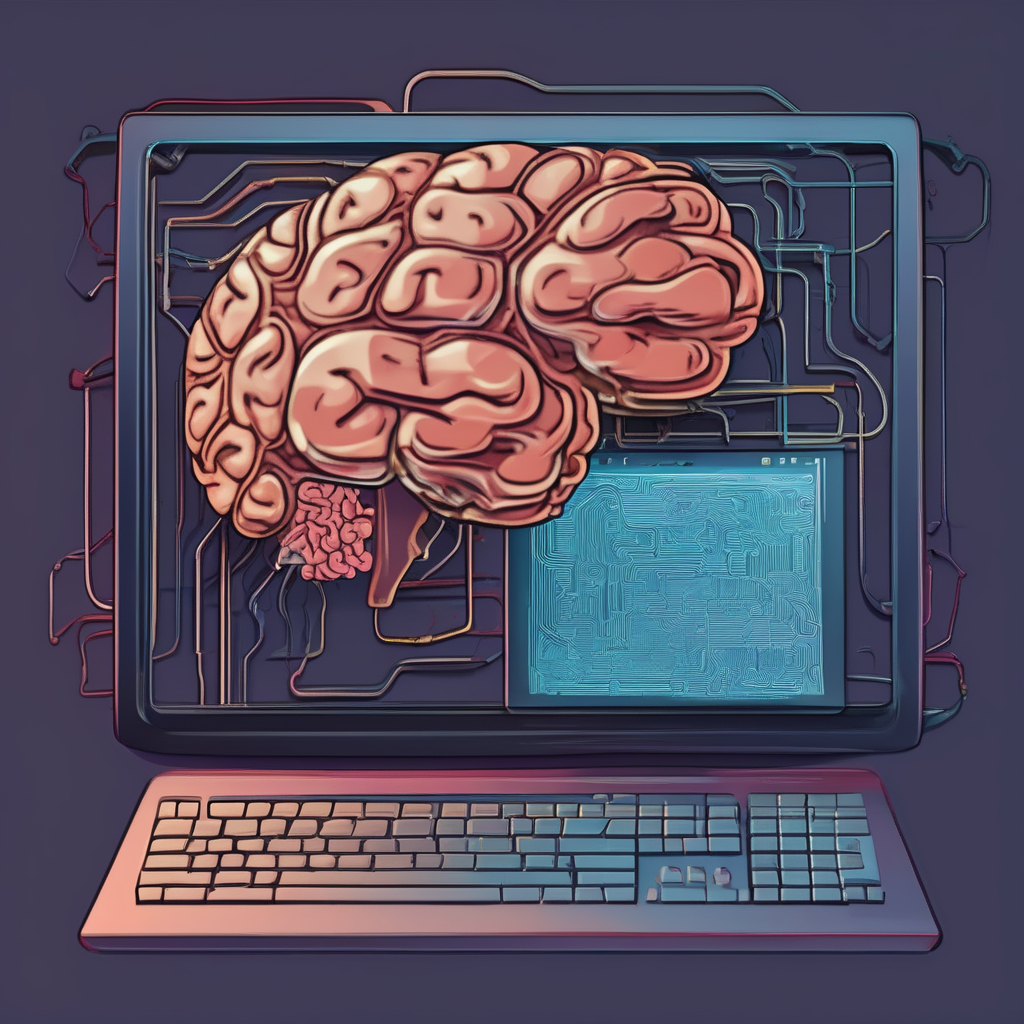}
        \caption{Lora}
    \end{subfigure}
    \hfill
    \begin{subfigure}{0.3\textwidth}
    \includegraphics[width=\linewidth]{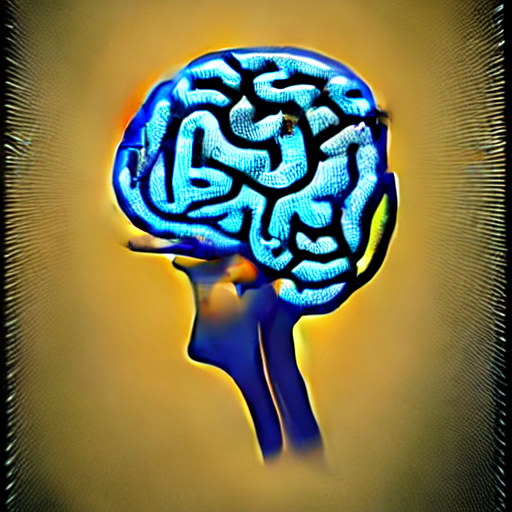}     
    \caption{SDXL (CFG=18)}
    \end{subfigure}
    \hfill
    \begin{subfigure}{0.3\textwidth}
        \includegraphics[width=\linewidth]{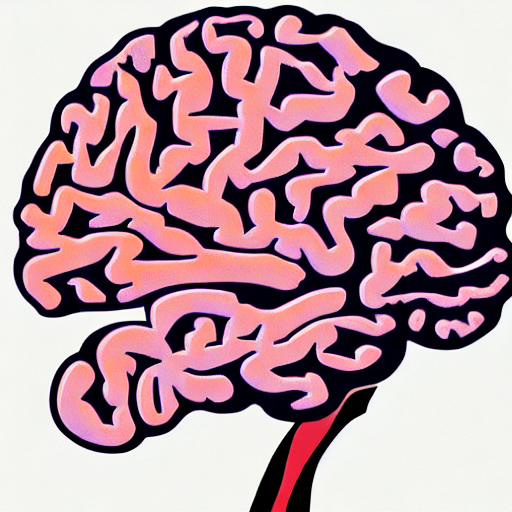}
        \caption{Attend and Excite}
    \end{subfigure}

    \vspace{0.3cm}
    
    \begin{subfigure}{0.3\textwidth}
        \includegraphics[width=\linewidth]{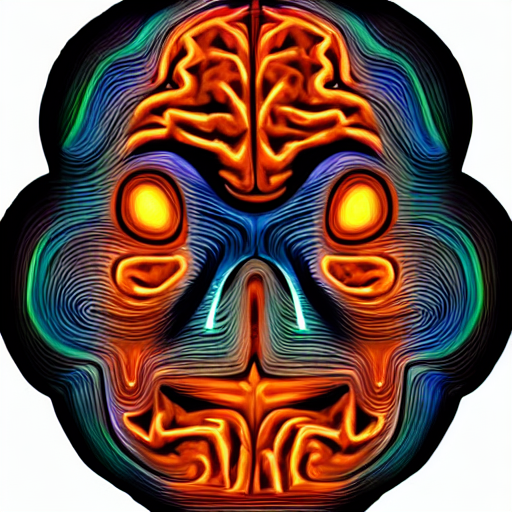}
        \caption{Controlnet}
    \end{subfigure}
    \hfill
    \begin{subfigure}{0.3\textwidth}
        \includegraphics[width=\linewidth]{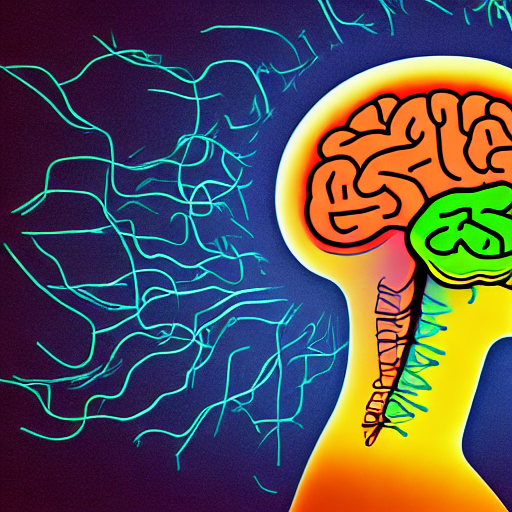}
        \caption{Dreambooth}
    \end{subfigure}
    \hfill
    \begin{subfigure}{0.3\textwidth}
        \includegraphics[width=\linewidth]{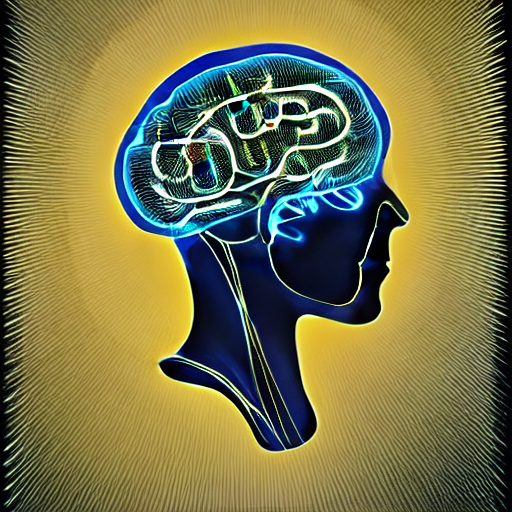}
        \caption{LPA "Ours"}
    \end{subfigure}

    \caption{Visual comparison for the prompt: \textit{“ A brain and a computer in neural style”}. LPA achieves consistent style and spatial structure.}
    \label{fig:cel_prompt_comparison3}
\end{figure}

\end{document}